\documentclass[conference]{IEEEtran}
\IEEEoverridecommandlockouts
\usepackage{cite}
\usepackage{amsmath,amssymb,amsfonts}
\usepackage{algorithmic}
\usepackage{graphicx}
\usepackage{textcomp}
\usepackage{xcolor}
\usepackage{verbatim} 
\usepackage{makecell} 
\usepackage{multirow}
\usepackage{CJKutf8}

\def\BibTeX{{\rm B\kern-.05em{\sc i\kern-.025em b}\kern-.08em
    T\kern-.1667em\lower.7ex\hbox{E}\kern-.125emX}}

\def\methodname{JPAVE}
\def\firstvariantname{JPAVE-GEN}
\def\secvariantname{JPAVE-CLS}
\begin{document}

\title{JPAVE: A Generation and Classification-based Model for Joint Product Attribute Prediction and Value Extraction}

\author{\IEEEauthorblockN{Zhongfen Deng}
\IEEEauthorblockA{\textit{Department of Computer Science}\\
\textit{University of Illinois Chicago}\\
Chicago, Illinois 60607\\
Email: zdeng21@uic.edu}
\and
\IEEEauthorblockN{Hao Peng}
\IEEEauthorblockA{\textit{School of Cyber Science and Technology}\\
\textit{Beihang University}\\
Beijing, China\\
Email: penghao@act.buaa.edu.cn}
\and
\IEEEauthorblockN{Tao Zhang}
\IEEEauthorblockA{\textit{Department of Computer Science}\\
\textit{University of Illinois Chicago}\\
Chicago, Illinois 60607\\
Email: tzhang90@uic.edu}
\and
\IEEEauthorblockN{Shuaiqi Liu}
\IEEEauthorblockA{\textit{Department of Computing}\\
\textit{The Hong Kong Polytechnic University}\\
HongKong, China\\
Email: cssqliu@comp.polyu.edu.hk}
\and
\IEEEauthorblockN{Wenting Zhao and Yibo Wang}
\IEEEauthorblockA{\textit{Department of Computer Science}, \\
\textit{University of Illinois Chicago}\\
Chicago, Illinois 60607\\
Email: wzhao41@uic.edu,
ywang633@uic.edu}
\and
\IEEEauthorblockN{Philip S. Yu}
\IEEEauthorblockA{\textit{Department of Computer Science}\\
\textit{University of Illinois Chicago}\\
Chicago, Illinois 60607\\
Email: psyu@uic.edu}
}

\maketitle

\begin{abstract}
Product attribute value extraction is an important task in e-Commerce which can help several downstream applications such as product search and recommendation. Most previous models handle this task using sequence labeling or question answering method which rely on the sequential position information of values in the product text and are vulnerable to data discrepancy between training and testing. This limits their generalization ability to real-world scenario in which each product can have multiple descriptions across various shopping platforms with different composition of text and style. They also have limited zero-shot ability to new values. In this paper, we propose a multi-task learning model with value generation/classification and attribute prediction called {\methodname} to predict values without the necessity of position information of values in the text. Furthermore, the copy mechanism in value generator and the value attention module in value classifier help our model address the data discrepancy issue by only focusing on the relevant part of input text and ignoring other information which causes the discrepancy issue such as sentence structure in the text. Besides, two variants of our model are designed for open-world and closed-world scenarios. In addition, copy mechanism introduced in the first variant based on value generation can improve its zero-shot ability for identifying unseen values. Experimental results on a public dataset demonstrate the superiority of our model compared with strong baselines and its generalization ability of predicting new values.

\end{abstract}


\section{Introduction}
The task of product attribute values extraction aims at identifying all the attribute values associated with a product by taking its text (e.g., product title or description) as input. 
This task has been studied for decades in e-Commerce due to its merits on improving 
customer's online shopping experience. It works as the fundamental task for many 
downstream applications such as product search and recommendation. Most previous research works treat it as a sequence labeling task and correspondingly many sequence labeling models are proposed such as \cite{zheng2018opentag,karamanolakis2020txtract,zhu-etal-2020-multimodal,yan2021adatag}.
OpenTag \cite{zheng2018opentag} introduces attention mechanism into to a BiLSTM-CRF sequence labeling method.  
TXtract \cite{karamanolakis2020txtract} improves attribute value extraction by utilizing hierarchical taxonomy of categories. 
Zhu \cite{zhu-etal-2020-multimodal} makes use of product images and propose a joint multimodal model to extract attribute values. \cite{wang-etal-2022-smartave} also propose a multimodal model following \cite{zhu-etal-2020-multimodal}.
AdaTag \cite{yan2021adatag} utilizes adaptive decoder for each attribute to motivate knowledge sharing among different attributes.
Some other existing works \cite{xu-etal-2019-scaling, wang2020learning} reduce attribute value extraction to answering some simple questions associating to the product text.  
The recent one \cite{roy2021attribute} proposes a novel attempt to generate values by fine-tuning pre-trained language models including GPT-2 \cite{radford2019language} and T5\cite{raffel2020exploring}.

Although these existing methods have achieved decent performance for attribute value extraction, they still have several drawbacks. The first drawback is \textbf{dependency of value position}. The existing sequence labeling and questions answering methods rely on the annotation of position information of attribute values in the product text, which is very costly in real-world applications as the annotation of position of values in a product text costs more than just annotating the values it has. In real world e-Commerce, most shopping platform or website may only have the weak-annotated attribute values provided by the merchant and no other annotation such as position of values in the text is available. It is almost impossible to train existing sequence labeling or question answering models in such cases.
The second one is that they \textbf{suffer from data shifting problem}. Sequence labeling methods are vulnerable to the data discrepancy between training and testing sets, which results in increasing adaptation studies \cite{han2019unsupervised,zhang-etal-2021-pdaln} to bridge the data gap. 
The training and testing data discrepancy is the different text's semantic and syntactic features. More specifically, the data resource and written style vary among datasets in most real-world scenarios, which leads to various sentence structures with different word distribution and usage. 
In the area of product attribute value extraction, the existing sequence labeling and question answering methods also suffer from the data shifting issue caused by the flexible sentence structure (e.g., expression order and composition) of the product description, which is common in e-Commerce. 
Because the same product can have multiple descriptions on various shopping website or platform with different composition and style (e.g., different expression order). These existing models trained on a dataset from one platform cannot be directly used to predict on another dataset from other platforms due to the data discrepancy between training and testing.
The third drawback is the \textbf{limited generalization ability to new attribute values}. New attribute values are not seen during the training stage, thus the model has no knowledge about them when encountering them in the testing phase.  
Although those existing methods mentioned before can be applied in open world scenario, their zero-shot ability of identifying new values is limited.

To address above mentioned limitations of existing methods for attribute value extraction, we propose a multi-task learning model called {\methodname} which is based on value generation/classification and attribute prediction. It has four major components which are text encoder, value generator, value classifier and attribute predictor. The first advantage of our proposed model is that it does not need the annotated position information of values in the product text. Because it uses a value generator or classifier to identify attribute values without regarding to the places where they appear. Furthermore, the copy mechanism introduced in the value generator and the value attention mechanism designed in the value classifier eliminate the negative effect of data inconsistency (e.g., different expression order and composition of product text) between training and testing. Because the copy mechanism or the value attention mechanism helps our model only focus on the relevant part of input text ignoring other information such as the structure of the sentence in the text.  
In addition, the copy mechanism designed in the value generator of our model also helps predict unseen values during testing phase which improves our model's zero-shot ability.
Besides the value generator, we also design a value classifier in our model to handle a scenario which has a pre-defined set of values in the dataset and only has annotated values for each product without their sequential position information. Our code is available for replication at: https://github.com/zhongfendeng/JPAVE.
In summary, our main contributions are as follows.
\begin{itemize}
\item To our best knowledge, this is the first work to design a multi-task learning model with value generation/classification and attribute prediction which is position-agnostic for identifying values in a product text. In other words, it does not rely on the sequential position information of values in the input product text. 
\item Our proposed model addresses data shifting problem in product attribute value extraction by only focusing on the relevant 
part of input text to identify attribute values (achieved by copy mechanism and value attention mechanism).
Therefore, other information in the text (e.g., the structure of sentence) which can cause data shifting does not affect our model's performance too much.
\item Our model has two variants designed for two different scenarios: open-world and closed-world scenario. The first variant {\firstvariantname} with value generation has a good generalization ability of predicting unseen values in test phase, the second one {\secvariantname} with value classification can be used in real-world scenario which only has labeled values for each product without any position information and has a pre-defined set of values.
\item Experimental results on a public dataset demonstrate the superiority of our model for attribute and value prediction including predicting both seen and unseen values.
\end{itemize}

\begin{figure}[h]
    \centering
    \includegraphics[scale=0.65]{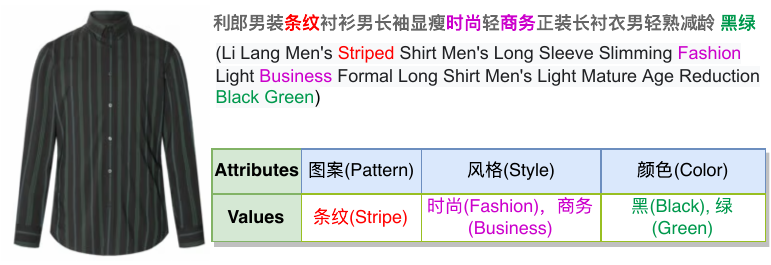}\vspace{-0.1in}
    \caption{An example showing product attribute prediction and value extraction.}\vspace{-0.25in}
    \label{fig:example}
\end{figure}

Although \cite{roy2021attribute} \footnote{The models in \cite{roy2021attribute} are not compared because of three reasons: 1) The models in \cite{roy2021attribute} are attribute dependent. 2) They need more input information (attribute), which will lead to an unfair comparison. 3) Their code and checkpoint are not publicly available.} is also a generative method for attribute value extraction, our proposed model is different from it in three perspectives. 
First, the problem definition is different. Specifically, \cite{roy2021attribute} take both the product text and its attribute as input to generate values, while our model only takes product text as input. We predict attributes and corresponding values simultaneously. Second, the model architecture is different. They fine-tune large-scale pre-trained models (GPT-2 and T5), while our model is light-weight and more efficient. The number of parameters of our model is only 12.3M, which is 10 and 18 times smaller than GPT2-small (124M) and T5-base (220M) in \cite{roy2021attribute}. Third, we use datasets in different languages. 
\cite{roy2021attribute} fine-tune GPT-2 and T5 on an English dataset \cite{xu-etal-2019-scaling}, while our model is trained on a Chinese dataset published by \cite{zhu-etal-2020-multimodal}.
Fig. \ref{fig:example} shows an example for joint product attribute prediction and value extraction. There are three attributes in this product and each attribute can have one or multiple values. Our model aims at predicting all the attributes and obtaining corresponding values for each attribute.

\section{Problem Formulation}
Given the textual description of a product, the goal is to predict what attributes exist for this product and to generate/classify all the attribute values appeared in the textual description. Assume that the textual product description is denoted as $T = (w_1, w_2,..., w_L)$, $L$ is the length of the textual description. The model aims at identifying multiple attributes denoted as $A = (a_1, a_2, ..., a_m)$ and attribute values denoted as $Y = (y_1, y_2, ..., y_n)$ for this product, where $m$ and $n$ are respectively the number of attributes and values this product has (each attribute may have multiple values, e.g., the attribute "color" has three values of black, white and red).  
Each product can have different numbers of attributes and attribute values. Suppose there are total $N_{attr}$ attributes and $N_{value}$ attribute values in a dataset. 
Our model  
aims at predicting multiple attributes out of the total $N_{attr}$ attributes for each product and generating/classifying all the corresponding values for each attribute in the product.

\section{Methodology}\label{sec:method}
\subsection{Overview}
\begin{figure*}[t]
    \centering
    \includegraphics[scale=0.70]{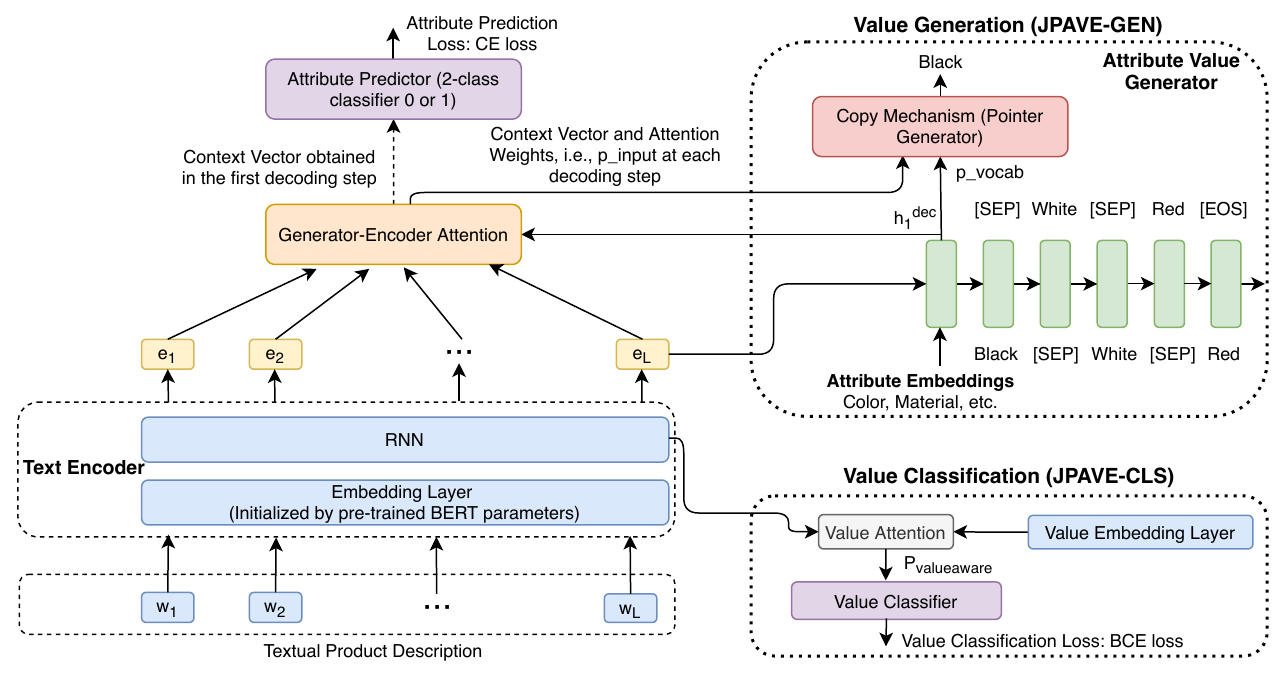}\vspace{-0.1in}
    \caption{The overall architecture of {\methodname}. [SEP] and [EOS] are special tokens.}\vspace{-0.25in}
    \label{fig:model_architecture}
\end{figure*}

The overall architecture of our proposed model {\methodname} is shown in Fig. \ref{fig:model_architecture}. 
There are two variations of {\methodname} which are {\firstvariantname} and {\secvariantname} (shown in the top-right and bottom-right of Fig. \ref{fig:model_architecture} respectively). These two variations share all components shown in the left half of Fig. \ref{fig:model_architecture} (including text encoder, attribute predictor and generator-encoder attention). The only difference between them is that {\firstvariantname} has an attribute value generator for value generation while {\secvariantname} adopts a value classifier for value prediction as the counterpart. We will describe the details of {\firstvariantname} and {\secvariantname} in following sections.

\subsection{{\firstvariantname}}
There are four components in {\firstvariantname} which are text encoder, generator-encoder attention module, attribute predictor and attribute value generator. Text encoder learns representations for each token in product text and the product representation. The generator-encoder attention module takes the hidden state of  
generator as the query to compute attention weights for each token in product text and then outputs a context vector by combining all token representations using the calculated attention weights. The attribute predictor takes the context vector generated in the first decoding step  
as input to predict attributes for each product. The attribute value generator uses the hidden state to predict a token from vocabulary at each decoding step by generating a probability distribution over the vocabulary. A copy mechanism \cite{see2017get} is used in the generator to help copy desired tokens into the generation of attribute values by considering the attention weights (i.e., another probability distribution) over the tokens in product text. This copy mechanism helps attribute value generator only focus on semantic meaning of each token in product text without regard to its position. Thus, no position information of attribute values is needed in {\firstvariantname}. Moreover, the copy mechanism also helps {\firstvariantname{}} handle the data shifting problem by ignoring information in product text such as the structure of sentence which causes such issue.

\subsubsection{Text Encoder}
The text encoder includes two parts, which are an embedding layer and a recurrent neural network (RNN). The embedding layer is initialized by using only the weights of embedding layer of the pre-trained BERT-base-chinese model \cite{devlin2018bert} (i.e., we do not use any other layers of BERT except the embedding layer, which reduces computational cost). The RNN is a Bi-GRU network. The working flow in the text encoder is as follows. Firstly, the embedding layer produces embeddings of all tokens in the input product description. Then, the RNN takes those token embeddings as input and 
produces contextual representation for each token. The last hidden state $e_L$ from the RNN is deemed as the product representation (i.e, the representation of the whole textual description). 
The text encoder in our model can be any encoder such as BERT encoder, Transformer encoder and so on. We adopt a RNN as the text encoder here for simplicity and memory cost consideration.

\subsubsection{Attribute Value Generator}
There can be multiple values for one attribute in a product in attribute value extraction. For instance, for a given product, the attribute "color" in this product may have values of black, white and red. The key point of transforming value extraction task into value generation is to prepare the desired target sequences for training the generator. Due to the goal of generating all values by one time for an attribute existing in a product, we compose the target value sequence for an attribute in a product by combining its multiple values using special tokens including [SEP] (separating each value) and [EOS] (indicating the end of composed sequence) as shown in the attribute value generator of Fig. \ref{fig:model_architecture}. 
During inference phase, the generated sequence for an attribute is parsed using [SEP] and [EOS] to obtain the multiple values generated.
In addition, attribute text containing semantic information is
utilized to obtain pre-trained attribute embedding to help attribute prediction and value generation. 
Suppose there are $N_{attr}$ attributes in the dataset, 
the tokens of each attribute are fed to the pre-trained BERT-base-chinese model, the CLS embedding outputted by BERT is taken as the embedding for the attribute. Finally, we can obtain an attribute embedding matrix denoted as $E_{attr} \in \mathbb{R}^{N_{attr}\times d_a}$, $d_a$ is the dimension of hidden state in BERT.

Based on the fact that all attribute values of a product come from its textual description, the copy mechanism  
can intuitively help produce correct attribute values. Thus the soft-gated pointer-generator copy mechanism is introduced in the generator to combine the probability distribution over the tokens of product description into the probability distribution over the vocabulary. The final combined probability distribution is used to predict the word of attribute values. 

It is worth to note that
any decoder structure can be utilized as the attribute value generator, we use a GRU-based decoder here for simplicity and memory cost consideration. The values of all attributes for the current product are generated independently at each decoding step. At the first decoding step, the pre-trained attribute embedding for the $i$-th attribute ($i=1,2,\cdot\cdot\cdot, N_{attr}$) is obtained from $E_{attr}$ and used as the decoder input, the last hidden state $e_L$ outputted by the text encoder is taken as the hidden state of previous GRU unit in the decoder. For the $j$-th decoding step of $i$-th attribute, the generator takes the embedding of the $j-1$-th word of its ground truth values (i.e., the composed target value sequence mentioned before) as input (as shown in Fig. \ref{fig:model_architecture}) and generates the hidden state of this step denoted as $h_{ij}^{dec}$,
then this hidden state is used to generate a probability distribution $p_{ij}^{vocab}$ over the vocabulary and a probability distribution $p_{ij}^{input}$ over the input textual tokens as shown in Eq. (\ref{eq:probs}), 
\begin{equation}\label{eq:probs}
    \small
    \begin{split}
    p_{ij}^{vocab} = Softmax(E(h_{ij}^{dec})^T) \in \mathbb{R}^{|V|}, \\
    p_{ij}^{input} = Softmax(H^{enc}(h_{ij}^{dec})^T) \in \mathbb{R}^{L},
    \end{split}
\end{equation}
where $E$ is the embedding matrix of vocabulary, $|V|$ is the size of vocabulary, $H^{enc} = [e_1; e_2; \cdot\cdot\cdot; e_L] \in \mathbb{R}^{L\times d_a}$ is the encoded representations of all tokens outputted by the text encoder.
The copy mechanism is implemented in Eq. (\ref{eq:copy}) to generate a weight $p_{ij}^{gen}$ to combine the above two distributions.
\begin{equation}\label{eq:copy}
    \small
    p_{ij}^{gen} = Sigmoid(W_{cm}[h_{ij}^{dec};w_{ij};c_{ij}]) \in \mathbb{R}^{1},
\end{equation}
where $W_{cm}$ is the trainable parameter of the copy mechanism, $w_{ij}$ is the embedding of input word (e.g., "black") at $j$-th decoding step for $i$-th attribute, $c_{ij} = p_{ij}^{input}\cdot H^{enc} \in \mathbb{R}^{d_a}$ is the context vector at $j$-th decoding step for $i$-th attribute which is the weighted sum of encoded representations of all input tokens.
The final combined probability distribution $p_{ij}^{final}$, calculated in Eq. (\ref{eq:p_final}), is used to predict the $j$-th word of values for $i$-th attribute.
\begin{equation}\label{eq:p_final}
    \small
    p_{ij}^{final} = p_{ij}^{gen}\times p_{ij}^{vocab}+(1-p_{ij}^{gen})\times p_{ij}^{input} \in \mathbb{R}^{|V|}.
\end{equation}

\subsubsection{Attribute Predictor}
Due to the relationship between each attribute and its corresponding values, besides the attribute value generator, we jointly train an attribute predictor to capture such relations. Specifically, for the $i$-th attribute, the first hidden state $h_{i1}^{dec}$ generated by the decoder is fed to the generator-encoder attention module to obtain the attentions weights over all input tokens and a context vector $c_{i1}$ which relates to the current attribute. The context vector $c_{i1}$ obtained at the first decoding step is then fed into the attribute predictor to do classification for the current $i$-th attribute as shown in Eq. (\ref{eq:attr_predictor}). There are two possible classes for each attribute  
which are \textit{exist} (which means the current $i$-th attribute exist in the given product) and \textit{none} (which means the current $i$-th attribute does not exist in the given product). The predictor will output one of the two classes for all attributes for the given product.
\begin{equation}\label{eq:attr_predictor}
    \small
    C_{i}^{attr} = Softmax(W_{c^{attr}}(c_{i1})^T)) \in \mathbb{R}^{2},
\end{equation}

During training process, cross entropy loss of this predictor shown in Eq. (\ref{eq:attr_pred_loss}) is used to optimize the model, where $a_i \in \mathbb{R}^{2}$ is the ground truth one-hot vector for $i$-th attribute.
\begin{equation}\label{eq:attr_pred_loss}
    \small
    L_{C}^{attr} =  \sum_{i=1}^{N_{attr}} -\log(C_{i}^{attr}(a_{i})^T).
\end{equation}

\subsubsection{Generator-Encoder Attention Module}
In order to capture the interaction between the generation of attribute values and the input textual product description, a generator-encoder attention module is utilized to help the generator pay more attention to the most relevant tokens in the input text at each decoding step. In this way, the generated words of attribute values can be derived from the input and thus improve the performance of seen value extraction as the values are exactly part of the input text, this also improves the generalization ability of our model on predicting unseen values.
The attention weights of different input tokens to the word generated at $j$-th decoding step for $i$-th attribute is the probability distribution over the input tokens $p_{ij}^{input}$ as shown in
Eq. (\ref{eq:probs}).

\subsubsection{Objective Function of {\firstvariantname}}

\noindent{\bf $\bullet$ \emph{Loss of Attribute Value Generator.}}
In attribute value generator, at each decoding step such as $j$-th step, there is a predicted word of value sequence for the current $i$-th attribute. Thus there is a cross-entropy loss for each decoding step which compares the predicted word with the ground truth word $y_{ij}$ at the current $j$-th decoding step. Therefore, we can obtain the loss of value generator by summing the losses from all decoding steps and attributes as shown in Eq. (\ref{eq:value_gen_loss}),
\begin{equation}\label{eq:value_gen_loss}
    \small
    L_{G}^{value} =  \sum_{i=1}^{N_{attr}}\sum_{j=1}^{|Y_{i}|} -\log(p_{ij}^{final}(y_{ij})^T),
\end{equation}
where $y_{ij} \in \mathbb{R}^{|V|}$ is the ground truth one-hot vector of the $j$-th token of target value sequence for $i$-th attribute, 
$|Y_{i}|$ is the number of tokens in the target value sequence of 
$i$-th attribute.

\noindent{\bf $\bullet$ \emph{Final Objective of {\firstvariantname}.}} The final objective function is the sum of the loss of attribute predictor and above stated generator loss: $L_{{\firstvariantname}}=L_{C}^{attr}+L_{G}^{value}$.

\subsection{{\secvariantname}}
The only different component in {\secvariantname} from {\firstvariantname} is the value classification module described below.
\subsubsection{Value Classification}
As shown in the bottom right of Fig. \ref{fig:model_architecture}, the value classification module has three parts which are value attention module, value embedding layer and value classifier. The value embedding layer is initialized by a pre-trained value embedding matrix $E_{value}$ obtained as follows. Firstly, the text of each value is obtained by combining the text of attribute and value using [SEP] token. For example, the text of value "black" is "color [SEP] black". Then such text is fed into the pre-trained BERT-base-chinese model and the CLS embedding outputted from BERT is adopted as the pre-trained value embedding. Finally, we can obtain a pre-trained value embedding matrix $E_{value} \in \mathbb{R}^{N_{value}\times d_a}$. 
The value attention module takes the representation of all tokens in product description $E_{txt} = (e_1,e_2,...,e_L) \in \mathbb{R}^{L\times d_a}$ as one input and the embeddings of all values $W_{v} \in \mathbb{R}^{N_{value}\times d_a}$ from value embedding layer as another input. It computes the attention weights $Attn_{v}$ over all values for the current product as shown in Eq. (\ref{eq:value_attn}) and generates a value-aware textual product representation $P_{valueaware}$ which is fed to the value classifier for value prediction. The value attention module only cares about the semantic representations of tokens in the product text without considering the position information of values in this text, which also eliminates the need of position information of values for {\secvariantname} like the copy mechanism does for {\firstvariantname{}}. In others words, both {\secvariantname} and {\firstvariantname} are positional invariant for identifying attribute values for a product. Furthermore, the value attention module also helps {\secvariantname} cope with data discrepancy between training and testing by focusing on relevant part of input text and ignoring other information such as sentence structure (e.g., different expression order) in the text.
\begin{equation}\label{eq:value_attn}
    \small
    \begin{split}
    Attn_{v} = Softmax((E_{txt}(W_{v})^T)^T) \in \mathbb{R}^{N_{value}\times L}, \\
    P_{valueaware} = mean(Attn_{v}E_{txt}) \in \mathbb{R}^{1\times d_a},
    \end{split}
\end{equation}
The value classification loss is the traditional binary cross entropy (BCE) loss shown as follows.
\begin{equation}\label{eq:value_clsloss}
    \small
    L_{C}^{value} = -\sum_{i=1}^{N} \sum_{j=1}^{N_{value}}[y_{ij}\log(y'_{ij})+(1-y_{ij})\log(1-y'_{ij})],
\end{equation}
where $N$ is the total number of training instances, $N_{value}$ is the total number of attribute values in the dataset, $y_{ij}$ and $y'_{ij}$ are the ground truth probability and predicted probability of $j$-th attribute value for $i$-th training instance respectively.

\subsubsection{Objective Function of {\secvariantname}}
The final objective function of {\secvariantname} is the sum of attribute prediction loss and value classification loss: $L_{{\secvariantname}} = L_{C}^{attr}+L_{C}^{value}$.

\section{Experiment and Analysis}
\subsection{Dataset}
We run experiments on a public available attribute value extraction dataset called MEPAVE\footnote{https://github.com/jd-aig/JAVE} released by \cite{zhu-etal-2020-multimodal}. It is a multimodal dataset which contains images for products. We only use the Chinese textual information (i.e., ignoring the images) of the products to train and evaluate our model and adopt the same split of training, validation and testing set as \cite{zhu-etal-2020-multimodal}. Table \ref{table:dataset_statistics} shows statistics of the dataset.
\begin{table}[h]
\small
\centering
\begin{tabular}{p{0.80cm}<{\centering} p{0.80cm}<{\centering} p{0.80cm}<{\centering} p{0.80cm}<{\centering} p{0.80cm}<{\centering}} 
 \hline
 \textbf{\#Attr} & \textbf{\#Value} & \textbf{\#Train} & \textbf{\#Val} & \textbf{\#Test} \\ [0.5ex] 
 \hline
 26 & 2,129 & 71,194 & 8,000 & 8,000 \\ [0.5ex]
\hline
\end{tabular}
\caption{Statistics of MEPAVE.  
\#Attr and \#Value are the total number of attributes and values in the dataset.}
\label{table:dataset_statistics}
\end{table}

\subsection{Evaluation Metrics}
We adopt the same evaluation metric F1 score as in JAVE \cite{zhu-etal-2020-multimodal} to evaluate our proposed model for attribute prediction and attribute value generation and classification. 
The F1 score  
is calculated 
in Eq. (\ref{eq:compute_f1})
, $Pred_{crt}, Pred_{total}, Gold_{total}$ are the number of correct predictions, total predictions, and ground truth attributes/values of all instances in the test set respectively. $P$ is precision, $R$ is recall. 
\begin{equation}\label{eq:compute_f1}
    \small
    \begin{split}
    P = Pred_{crt}/Pred_{total}, R =Pred_{crt}/Gold_{total}, \\ F1 =2*P*R/(P+R).
    \end{split}
\end{equation}

For attribute prediction, our model is evaluated in the same way as JAVE. For value prediction, both our model and JAVE are evaluated by F1 score calculated in Eq. (\ref{eq:compute_f1}). The only difference is that our model extracts predicted values for a product in a different way. JAVE uses the predicted BIO tags to extract a chunk of product text as a value, while our model parses the generated text by using special token [SEP] and [EOS] to get values for a product. The ground truth values of each product are the same for any model.
Thus, the results of our model are comparable to JAVE and all other baselines. 
We use Exact Match \cite{rajpurkar2016squad} to calculate the scores.

\noindent{\bf $\bullet$ \emph{Additional Evaluation Metrics for Value Generation.}}
In addition, we also use other metrics similar to the metrics in \cite{wu-etal-2019-transferable} including Joint ACC, Instance-level ACC and Joint F1 to evaluate the value generation performance of our model. 

For Joint ACC (\textbf{JACC}), if all the ground truth values for the current instance 
are predicted correctly, then the number of correctly predicted instance $N_{pred}$ will be increased by one, the Joint ACC is the ratio between the number of completely and correctly predicted instance and the total number of instance $N_{total}$: $JACC = N_{pred}/N_{total}$.

For the Instance-level ACC (\textbf{IACC}), the accuracy for each instance, such as the $i$-th instance, is $ins_{acc} = N_{predvalue}^{ins}/N_{totalvalue}^{ins}$, where $N_{predvalue}^{ins}$ and $N_{totalvalue}^{ins}$ are the number of correctly predicted values and the total number of values for the current instance respectively. The final Instance-level ACC is the average of $ins_{acc}$ from all instances: $IACC= \frac{1}{N_{total}}\sum_{i=1}^{N_{total}} ins_{acc}^{i}$.

For Joint F1 (\textbf{JF1}), it is calculated in the similar way as IACC. Firstly, for each instance, we compute the Micro-F1 score $ins_{f1}$ by using the ground truth attribute values and the generated ones, then the Joint F1 is the average of $ins_{f1}$ from all instances: $JF1 = \frac{1}{N_{total}} \sum_{i=1}^{N_{total}} ins_{f1}^{i}$.

Apart from the standard evaluation metric of F1 score for both attribute prediction and value generation, these three metrics can further demonstrate the model's ability of identifying all attribute values mentioned in the product as a whole.

\subsection{Baselines}
We compare our model with the model \textbf{JAVE} and \textbf{M-JAVE} proposed by \cite{zhu-etal-2020-multimodal} and the \textbf{SMARTAVE} model proposed by \cite{wang-etal-2022-smartave}. The baselines used in \cite{zhu-etal-2020-multimodal} including \textbf{RNN-LSTM} \cite{hakkanitur16_interspeech}, \textbf{Attn-BiRNN} \cite{liu16c_interspeech}, \textbf{Slot- Gated} \cite{goo-etal-2018-slot}, \textbf{Joint-BERT} \cite{chen2019bert} and \textbf{ScalingUp} \cite{xu-etal-2019-scaling} are also compared. The first four baselines are models for intent classification and slot filling tasks. The last baseline \textbf{ScalingUp} is a sequence labeling model with attention mechanism, BiLSTM and CRF. \textbf{JAVE} is trained without product images, while \textbf{M-JAVE} is a multimodal model trained with product images. \textbf{SMARTAVE} is also a multimodal model which has a variation trained only with text called \textbf{SMARTAVE-text}. Two baselines (\textbf{AVEQA} \cite{wang2020learning} and \textbf{MAVEQA}\cite{yang2022mave}) used in \cite{wang-etal-2022-smartave} are also compared with our model.
\textbf{AVEQA} \cite{wang2020learning} is a question answering method for attribute value extraction. It encodes both attribute and product text to obtain the value. \textbf{MAVEQA}\cite{yang2022mave} uses a new encoder to encode structure and longer input sequences.

\subsection{Experimental Setting}
All experiments are conducted on a server with a single NVIDIA GeForce RTX 3090 graphics card with 24G memory.
For the hyper-parameters, the maximum length of token sequence of product description is set to 46 as in \cite{zhu-etal-2020-multimodal}, the dimensions of text, attribute and value embedding are 768, the size of text encoder hidden state and decoder hidden state are 384 and 768 respectively. The maximum length of target and generated value sequence  
are 10. The batch size for training, validation and testing are 64, 16 and 16 respectively.

\subsection{Main Results and Analysis}
The experimental results of our model and baselines are shown in Table \ref{table:exp_results}. The results of baselines are directly quoted from \cite{zhu-etal-2020-multimodal} and \cite{wang-etal-2022-smartave}. From this table, we can see that our model outperforms all the baselines by a large margin. Specifically, {\firstvariantname} and {\secvariantname} achieve improvements of 11.8\%, 15.0\% and 11.8\%, 9.6\% respectively on attribute prediction and value prediction compared with JAVE. Furthermore, they also perform much better than the multimodal model M-JAVE which utilizes the product image information besides the textual product description by improvements of 8.5\%, 7.2\% and 8.5\%, 6.6\% on attribute and value prediction respectively. When compared with the most recent multimodal model SMARTAVE and its variant SMARTAVE-text which only uses product text information, our model still obtains significant improvements.
This demonstrates the effectiveness and superiority of our model on attribute value extraction task.  
{\firstvariantname} performs better than {\secvariantname} on both tasks (i.e., attribute and value prediction), especially for value extraction. This may be the reason that generation model generates desired word of values by focusing more on the input text  
which makes it more likely to produce the correct values. While the classification model tries to predict the correct values out of more than two thousands of values (2,129 values in total) most of which are irrelevant to the current product. This makes it harder for 
{\secvariantname} to predict the correct values for the current product. However, {\secvariantname} still obtains decent improvements compared with the sequence labeling baselines on both tasks.

\begin{table}[t]
\small
\centering
\begin{tabular}{p{3.2cm}<{} | p{1.4cm}<{\centering} p{1.4cm}<{\centering}} 
 \hline
	\multirow{1}{*}{\textbf{Model}} & \textbf{Attribute} & \textbf{Value} \\ [0.5ex] 
 \hline
 \textbf{RNN-LSTM} \cite{hakkanitur16_interspeech} & 85.76 & 82.92 \\ [0.5ex]
 \textbf{Attn-BiRNN} \cite{liu16c_interspeech} & 86.10 & 83.28 \\ [0.5ex]
 \textbf{Slot-Gated} \cite{goo-etal-2018-slot} & 86.70 & 83.35 \\ [0.5ex]
 \textbf{Joint-BERT} \cite{chen2019bert} & 86.93 & 83.73 \\ [0.5ex]
 \textbf{ScalingUp} \cite{xu-etal-2019-scaling} & - & 77.12 \\ [0.5ex]
 \textbf{AVEQA} \cite{wang2020learning} & - & 89.15 \\ [0.5ex]
 \textbf{MAVEQA} \cite{yang2022mave} & - & 88.71 \\ [0.5ex]
 \textbf{SMARTAVE-text} \cite{wang-etal-2022-smartave} & - & 89.21 \\ [0.5ex]
 \textbf{SMARTAVE} \cite{wang-etal-2022-smartave} & - & 91.52 \\ [0.5ex]
 \textbf{JAVE} \cite{zhu-etal-2020-multimodal} & 87.98 & 84.78 \\ [0.5ex]
 \textbf{M-JAVE} \cite{zhu-etal-2020-multimodal} & 90.69 & 87.17 \\ [0.5ex]
 \hline
 \textbf{{\firstvariantname}} (Ours) & \textbf{98.40} & \textbf{97.51} \\ [0.5ex]
 \textbf{{\secvariantname}} (Ours) & \textbf{98.39} & \textbf{92.88} \\ [0.5ex]
\hline
\end{tabular}
\caption{Main results (F1 score \%) of attribute prediction and value extraction of our model and baselines. We only compare the results of BERT-based \textbf{JAVE} and \textbf{M-JAVE} model proposed in \cite{zhu-etal-2020-multimodal} because the LSTM-based models perform worse than the BERT-based models. "-" means the corresponding baseline only extracts values.
}
\label{table:exp_results}
\end{table}

\begin{figure*}[t]
    \centering
    \includegraphics[scale=0.55]{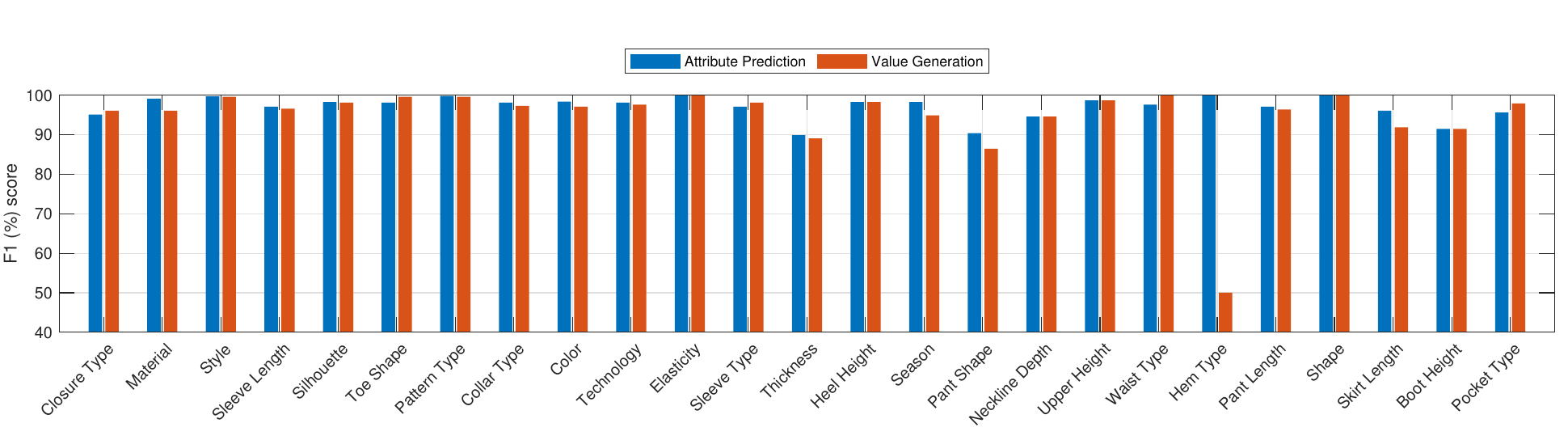}\vspace{-0.1in}
    \caption{The F1 (\%) score for each attribute in MEPAVE dataset obtained by attribute predictor and attribute value generator in {\firstvariantname}.}\vspace{-0.25in}
    \label{fig:results_eachattr}
\end{figure*}

To further verify the superiority of our model, we analyze the F1 score on each attribute obtained by {\firstvariantname} on both attribute and value prediction. The results are shown in Fig. \ref{fig:results_eachattr}. One can see that the F1 scores of most attributes on both attribute and value prediction exceed 90\% and almost close to 100\%. This further demonstrates that our model can correctly predict almost all existing attributes and their corresponding values for a given product.

\noindent{\bf $\bullet$ \emph{Performance of {\methodname} in two scenarios: open-world and closed-world scenario.}}
To demonstrates the effectiveness of our model for two different real-world scenarios, we show the performance results of the two variants on seen and unseen values respectively in Table \ref{table:ab_twoscenarios}. One can see that {\firstvariantname} performs well on both seen and unseen values, which makes it appropriate for open-world scenario that constantly has emerging new values. {\firstvariantname} w/o Copy performing much worse especially on unseen values validates the effect of copy mechanism on helping our model generalize well to new values. Although {\secvariantname} cannot predict unseen values, it performs well on the seen values compared with all the baselines.  
Thus it is suitable for the closed-world scenario which only has annotated values without position information and has a pre-defined set of values. 
\begin{table}[h]
\small
\centering
\begin{tabular}{p{3.1cm}<{} | p{0.9cm}<{\centering} p{1.3cm}<{\centering} p{1.6cm}<{\centering}} 
 \hline
	\multirow{1}{*}{\textbf{Model}} & \textbf{All} & \textbf{SeenValues} & \textbf{UnseenValues} \\ [0.5ex] 
 \hline
 {\firstvariantname} & 97.51 & 97.65 & 83.15 \\ [0.5ex]
 {\firstvariantname} w/o Copy & 94.86 & 95.28 & 36.22 \\ [0.5ex]
 \hline
 {\secvariantname} & 92.88 & 93.37 & 00.00 \\ [0.5ex]
\hline
\end{tabular}
\caption{F1 scores of {\methodname} for value prediction in closed and open world scenario.}
\label{table:ab_twoscenarios}
\end{table}

\begin{table}[h]
\small
\centering
\begin{tabular}{p{1.8cm}<{} | p{1.4cm}<{\centering} p{1.0cm}<{\centering} p{1.0cm}<{\centering}} 
 \hline
	\multirow{1}{*}{\textbf{Attribute}} & \textbf{Precision} & \textbf{Recall} & \textbf{F1} \\ [0.5ex] 
 \hline
 Material & 100.00 & 66.67 & 80.00 \\ [0.5ex]
 Style & 100.00 & 87.50 & 93.33 \\ [0.5ex]
 Pattern Type & 100.00 & 90.00 & 94.74 \\ [0.5ex]
 Color & 100.00 & 80.56 & 89.23 \\ [0.5ex]
 Technology & 100.00 & 60.00 & 75.00 \\ [0.5ex]
 \hline
 \textbf{Overall} & 100.00 & 71.15 & 83.15 \\ [0.5ex]
\hline
\end{tabular}
\caption{Zero-shot results of {\firstvariantname}.}
\label{table:ab_unseenvalues}
\end{table}

\noindent{\bf $\bullet$ \emph{Zero-Shot Ability of {\firstvariantname}.}}
To further demonstrate the generalization ability of {\firstvariantname} on predicting unseen values, we show the performance of {\firstvariantname} on all unseen values in the test set belonging to different attribute in Table \ref{table:ab_unseenvalues}.  
There are total 99 unseen values in the test set. 
We group these unseen values into their corresponding attribute, the zero-shot results of {\firstvariantname} on different attribute 
are
shown in Table \ref{table:ab_unseenvalues}. One can see that {\firstvariantname} obtains a good zero-shot ability of correctly generating unseen values belonging to different category of attributes. Especially, it achieves a high zero-shot performance with 94.74\% and 93.33\% F1 score on the attribute of pattern type and style respectively. 
This excellent zero-shot ability of {\firstvariantname} is very helpful in real-world applications as new products emerge very often in e-Commerce. The 100\% precision of all unseen values demonstrates the effectiveness of copy mechanism on correctly identifying unseen values from input product text.

\noindent{\bf $\bullet$ \emph{Position-Agnostic of {\methodname} and Its Robustness to Data Shifting.}}
To verify that our model is invariant to the position of attribute values in the input text, and that it can cope with the data shifting problem caused by the flexible sentence structure of the product text in attribute value extraction (i.e., it is robust to the data inconsistency between training and testing), we conduct experiments for our model on one more testing set generated by ourselves. Specifically, we first create a permuted test set by randomly permuting the order of words in the input product text for each testing instance (The order of words in each attribute value is not changed).
Then our trained model is directly tested on this permuted test set besides the normal one. The results of our model on these two test sets (i.e., both permuted and original test set) are shown in Table \ref{table:ps_results}. We can see that although the F1 scores of attribute and value prediction on the permuted test set drop a little bit, our model still achieves very good results which are better than all the baselines tested on the original test set without perturbation. This demonstrates that our model is agnostic and insensitive to the position of attribute values in the input text for predicting both attributes and their corresponding values. It also validates that our model is free from data shifting caused by flexible expression order and composition of product text in product attribute value extraction. These properties of our model
can be very helpful in real-world applications in which one product may have multiple different composition of title and textual descriptions across different shopping platform or website. They can also help reduce the annotation workload by eliminating the need of tagging every token in a sequence.
\begin{table}[h]
\small
\centering
\begin{tabular}{p{1.82cm}<{} |p{0.9cm}<{\centering} p{0.9cm}<{\centering} |p{0.9cm}<{\centering} p{0.9cm}<{\centering}} 
 \hline
	\multirow{2}{*}{\textbf{Model}} & \multicolumn{2}{c|}{Original Test Set} & \multicolumn{2}{c}{Permuted Test Set} \\ 
	\cline{2-5}
	& \textbf{Attr} & \textbf{Value} & \textbf{Attr} & \textbf{Value} \\ [0.5ex] 
 \hline
 \textbf{{\firstvariantname}} & 98.40 & 97.51 &97.61 & 94.75 \\ [0.5ex]
 \textbf{{\secvariantname}} & 98.39 & 92.88 & 98.24 & 90.07 \\ [0.5ex]
\hline
\end{tabular}
\caption{F1 results of {\methodname} for attribute and value prediction on two test sets (i.e., the original and permuted test set).}
\label{table:ps_results}
\end{table}

\subsection{Ablation Study}
\subsubsection{Ablation Study for {\firstvariantname}}
To further verify the effect of copy mechanism in attribute value generator, the joint learning and the pre-trained attribute embeddings, we conduct ablation studies for {\firstvariantname} whose results are shown in Table \ref{table:ab_results}.
Four variants are as follows.

\noindent{\bf $\bullet$ \emph{{\firstvariantname} w/o Copy.}}
It removes the copy mechanism in value generator and predicts the word of values at each step by only using the probability distribution over the vocabulary.

\noindent{\bf $\bullet$ \emph{{\firstvariantname} w/o APred.}} 
It removes the attribute predictor.  
The loss of this variant is $L_{G}^{value}$.

\noindent{\bf $\bullet$ \emph{{\firstvariantname} frz-AEmb.}}
The attribute embeddings in value generator are initialized by pre-trained attribute embedding matrix $E_{attr}$, but they are frozen during the training process.

\noindent{\bf $\bullet$ \emph{{\firstvariantname} rnd-AEmb.}}
The attribute embeddings in the generator are randomly initialized and updated during training by back propagation.

From Table \ref{table:ab_results}, one can see that the variant \textit{{\firstvariantname} w/o Copy} performs much worse than {\firstvariantname} on value generation, which demonstrates that the copy mechanism in attribute value generator can get tokens from input text and thus helps generate word of values appeared in the input text and improves the performance of value generation. 
\textit{{\firstvariantname} w/o APred} achieves slightly better results on value generation than {\firstvariantname}. This maybe because 
the optimization directions of the two losses in {\firstvariantname} are a little divergent from each other (i.e., not completely consistent with each other) which makes the loss of value generation not minimized as well as the only one loss in \textit{{\firstvariantname} w/o APred}. However, {\firstvariantname} obtains almost the same performance as this variant on value generation and performs much better on attribute prediction which demonstrates the versatility of {\firstvariantname} and that the joint training is effective for both tasks.
The variant with frozen pre-trained attribute embeddings and the one with randomly initialized embeddings performance a little worse than {\firstvariantname} on value generation, which demonstrates the usefulness (although not very big) of pre-trained attribute embeddings obtained by using the text information of attribute.

\begin{table}[h]
\small
\centering
\begin{tabular}{p{3.29cm}<{} | p{0.6cm}<{\centering} p{0.6cm}<{\centering} p{0.6cm}<{\centering} p{0.6cm}<{\centering} p{0.6cm}<{\centering}} 
 \hline
	\multirow{1}{*}{\textbf{Model}} & \textbf{Attr} & \textbf{Value} & \textbf{JACC} & \textbf{IACC} & \textbf{JF1} \\ [0.5ex] 
 \hline
 {\firstvariantname} w/o Copy & 98.40 & 94.86 & 91.54 & 99.61 & 95.10 \\ [0.5ex]
 {\firstvariantname} w/o APred & 20.14 & \textbf{97.77} & \textbf{94.55} & \textbf{99.77} & \textbf{97.38} \\ [0.5ex]
 {\firstvariantname} frz-AEmb & \textbf{98.46} & 97.43 & 93.59 & 99.72 & 96.75 \\ [0.5ex]
 {\firstvariantname} rnd-AEmb & 98.45 & 97.38 & 93.94 & 99.74 & 96.97 \\ [0.5ex]
 \textbf{{\firstvariantname}} & 98.40 & \textbf{\textit{97.51}} & \textbf{\textit{94.22}} & \textbf{\textit{99.75}} & \textbf{\textit{97.15}} \\ [0.5ex]
\hline
\end{tabular}
\caption{Ablation study results for {\firstvariantname}. w/o means without the corresponding component.}
\label{table:ab_results}
\end{table}

\subsubsection{Ablation Study for {\secvariantname}}
To demonstrate the helpfulness of making use of text information of values for 
value classification, we also conduct ablation studies for 
{\secvariantname} whose results are shown in Table \ref{table:ab_results_1}. There are two variants of {\secvariantname} as follows.

\noindent{\bf $\bullet$ \emph{{\secvariantname} rnd-ValueEmb.}}
It does not make use of text information of values. The value embedding layer is randomly initialized and updated during training by back propagation.

\noindent{\bf $\bullet$ \emph{{\secvariantname} freeze-ValueEmb.}}
The value embedding layer in this variant is initialized by pre-trained value embeddings obtained from pre-trained BERT model which uses the text information of values. However, it is frozen during training.

Form Table \ref{table:ab_results_1}, we can see that the variant with randomly initialized value embedding layer and the one with frozen value embeddings perform 
worse than {\secvariantname} on attribute prediction and value prediction respectively. This indicates that pre-trained value embedding obtained by making use of the text of one value and the attribute it belongs to gives a little bit help to {\secvariantname} for classifying attributes and values.

\begin{table}[ht]
\small
\centering
\begin{tabular}{p{4.02cm}<{} | p{1.2cm}<{\centering} p{1.2cm}<{\centering}} 
 \hline
	\multirow{1}{*}{\textbf{Model}} & \textbf{Attr} & \textbf{Value} \\ [0.5ex] 
 \hline
 {\secvariantname} rnd-ValueEmb & 98.12 & 92.88 \\ [0.5ex]
 {\secvariantname} freeze-ValueEmb & \textbf{98.48} & 90.05 \\ [0.5ex]
 \textbf{{\secvariantname}} & 98.39 & \textbf{92.88} \\ [0.5ex]
\hline
\end{tabular}
\caption{Ablation study results for {\secvariantname}.}
\label{table:ab_results_1}
\end{table}

\begin{table*}[ht]
\small
\centering
\begin{tabular}{p{0.01cm}<{}|p{7.50cm}<{} |p{2.9cm}<{\centering} |p{3.1cm}<{\centering}|p{2.6cm}<{\centering}} 
 \hline
	& \multirow{1}{*}{\textbf{Product Text}} & \textbf{Ground-truth Values} & \textbf{Predicted Values ({\firstvariantname})} & \textbf{Predicted Values ({\secvariantname})} \\ [0.5ex] 
 \hline
 \multirow{2}{*}{1} & \begin{CJK*}{UTF8}{gbsn}\underline{复古}感\underline{格纹}长裤结合厚实的\underline{毛呢}面料\end{CJK*} (\textbf{Plaid} pants of \textbf{old-fashion} style made with material of thick \textbf{woolen} fabric) & \multirow{2}{*}{Old-fashion, Plaid, Woolen} & Woolen, Old-fashion, Plaid & Old-fashion, Plaid, Long-pants \\ [0.5ex]
 \cline{4-5}
 & \begin{CJK*}{UTF8}{gbsn}感长\underline{格纹}裤\underline{复古}结合厚\underline{毛呢}实的面料\end{CJK*} (\textit{permuted}) (Of style pants \textbf{plaid} \textbf{old-fashion} made with thick material of \textbf{woolen} fabric) &  & Woolen, Old-fashion, Plaid & Old-fashion, Plaid \\ [0.5ex]
 \hline
 \multirow{2}{*}{2} & \begin{CJK*}{UTF8}{gbsn}雪豹\underline{秋冬} \underline{真皮}皮衣男\underline{绵羊皮}男士\underline{立领}皮夹克\end{CJK*} (Men's \textbf{Stand Collar} \textbf{Genuine Leather} Jacket of Snow Leopard made with \textbf{sheepskin} \textbf{leather} for \textbf{Autumn and Winter}) & \multirow{2}{2.9cm}{Autumn and Winter, \\ Genuine Leather, Sheepskin, \\ Leather, Stand Collar} &  Genuine Leather, Sheepskin, Stand Collar, Autumn and Winter & Leather, Autumn and Winter \\ [0.5ex]
 \cline{4-5}
 & \begin{CJK*}{UTF8}{gbsn}雪\underline{秋冬} \underline{绵羊皮}豹皮衣\underline{真皮}男男\underline{立领}皮士夹克\end{CJK*} (\textit{permuted}) (Men's Jacket \textbf{Genuine Leather} \textbf{Stand Collar} of Snow \textbf{Autumn and Winter} \textbf{sheepskin} \textbf{leather} Leopard made with) &  & Genuine Leather, Stand Collar, Autumn and Winter & Leather \\ [0.5ex]
 \hline
 \multirow{2}{*}{3} & \begin{CJK*}{UTF8}{gbsn}\underline{红色}的手提和包身的\underline{拼接}线条形成色彩呼应\end{CJK*} (The \textbf{red} handle and the \textbf{splicing} color of the body of the bag form a color echo) & \multirow{2}{2.9cm}{Red, Splicing} & Splicing, Red & Splicing, Red \\ [0.5ex]
 \cline{4-5}
 & \begin{CJK*}{UTF8}{gbsn}的\underline{拼接}手提和包身的线条形成色\underline{红色}彩呼应\end{CJK*} (\textit{permuted}) (The \textbf{splicing} handle and the body of the bag form a color \textbf{red} echo) &  & Splicing, Red & Splicing, Red \\ [0.5ex]
 
\hline
\end{tabular}
\caption{Case study results for {\methodname}. Each example has two versions of product text, the top version is the original one without perturbation, the bottom one is the permuted text (the text in parenthesis are the English translation of the Chinese product descriptions). Underlined text in the Chinese product description and the corresponding bold text in the English translation are the desired attribute values we want to extract.}
\label{table:casestudy_result}
\end{table*}

\subsection{Case Study}
To further demonstrate the position-insensitivity of our model and its ability of coping with data discrepancy caused by flexible expression order and composition of product text between training and testing, we show the prediction results of our model on  
several examples with two different compositions (i.e., the original one and the permuted one) in Table \ref{table:casestudy_result}. For each example, the top one of text is the original composition, the below one is generated by randomly permuting the order of words in text (the order of words in each attribute value is still kept the same as the original one). From this table, one can see that our model can still correctly predict the attribute values existing in each product text even if the order of words and sentence structure in the text is severely perturbed, such as the first and third example in the table.
Although in the second example, our model misses some values for the permuted product text, it (especially {\firstvariantname}) still predicts most of the informative values. This further demonstrates the position-insensitivity and robustness to data shifting of our model for real-world applications which may have different description text for the same product. This is because our model does not rely on the positional information of values in the product text and thus can predict the values no matter where they appear in the text. And our model ignores other information in product text which can cause data shifting such as the structure of sentence.

\section{Related Work}
\subsection{Product Attribute Value Extraction} 
Earlier models for product attribute value extraction are rule-based extraction methods \cite{gopalakrishnan2012matching,vandic2012faceted}. 
Later on, \cite{more2016attribute} treats it as a sequence labeling task and many models are designed \cite{sharma2016automatic,ramos2019attribute,Embar2021DiffXtractJD}. 
\cite{zheng2018opentag} propose a Bidirectional LSTM-CRF model with attention mechanism. 
\cite{xu-etal-2019-scaling, wang2020learning} formulate it as a question answering task.
\cite{mehta2021latex} propose a scalable framework for numeric value extraction. 
\cite{karamanolakis2020txtract} design a taxonomy-aware sequence labeling model which uses
hierarchical taxonomy of product categories.
\cite{yan2021adatag} propose AdaTag which allows knowledge sharing across different attributes using 
adaptive decoders.
There are also several datasets in different languages proposed for the task of product attribute value extraction in recent years. \cite{zhu-etal-2020-multimodal} propose a Chinese multimodal product attribute value dataset. \cite{yang2022mave} create a large scale dataset in English from the Amazon Review Dataset \cite{ni2019justifying} by adding annotations of attribute values for the products. 
\cite{xu-etal-2019-scaling} collect an English dataset from the category of Sports \& Entertainment on AliExpress, on which \cite{roy2021attribute} fine-tune GPT-2 and T5 to generate values by taking both the product text and its attribute as input.
We mainly focus on product attribute value extraction on Chinese dataset in this paper. We will explore cross-lingual product attribute value extraction in the future as different languages have different characteristics.

\subsection{Dialog State Tracking}
Dialog State Tracking (DST) is a fundamental task in dialog system.
The goal of DST is to keep track of user's belief states during the course of conversation. Belief states are usually represented as multiple pairs of domain-slot and value, such as (restaurant-price, cheap), (restaurant-area, north) and so on.
There are many DST models proposed in recent years, 
TRADE \cite{wu-etal-2019-transferable} is one of the most important models for multi-domain DST from which we get inspiration for our proposed model.

\subsection{Multi-Label Text Classification} 
We briefly introduce some works in multi-label text classification as {\methodname} includes a value classification module. 
There are mainly two groups of works in this area which are local models and global models.
Local models usually build a classifier for each label or labels in the same level of label taxonomy, such as \cite{wehrmann2018hierarchical,banerjee2019hierarchical,huang2019hierarchical}.
Global models build one classifier for all labels \cite{johnson2015effective,mao2019hierarchical,wu2019learning,peng2018large,peng2019hierarchical,zhou2020hierarchy,deng2021htcinfomax,deng2022aesmnsmlc}.  
Also, attention-based models are popular and designed for text classification in very recent years \cite{you2019attentionxml,chang2020taming,deng-etal-2020-hierarchical}.

\section{Conclusion}
In this paper, we propose a multi-task learning model with value generation/classification and attribute prediction to overcome previous models' dependency of position information of values in the text and their vulnerability to the data shifting problem caused by the flexible sentence structure (e.g., expression order and composition) of the product text in attribute value extraction (i.e., the data discrepancy between training and testing sets). Two variants of our model are designed for open-world and closed-world scenario. Copy mechanism is utilized in the variant based on value generation to improve its generalization ability of predicting new attribute values. Experimental results on a public dataset demonstrate the superiority of our model and its good zero-shot ability to unseen attribute values.

\section*{Acknowledgment}
We thank the reviewers for their comments and feedback. This work is supported by National Key R\&D Program of China through grant 2021YFB1714800 and the Fundamental Research Funds for the Central Universities and is supported in part by NSF under grant III-2106758.


\vspace{12pt}


\begin{thebibliography}{00}
\bibitem{banerjee2019hierarchical}
Siddhartha Banerjee, Cem Akkaya, Francisco Perez Sorrosal, and Kostas Tsioutsiouliklis. 2019. Hierarchical transfer learning for multi-label text classification. In Proceedings of the 57th ACL, pages 6295–6300.
\bibitem{chang2020taming}
Wei-Cheng Chang, Hsiang-Fu Yu, Kai Zhong, Yiming Yang, and Inderjit S Dhillon. 2020. Taming pre-trained transformers for extreme multi-label text classification. In Proceedings of the 26th ACM SIGKDD, pages 3163–3171.
\bibitem{chen2019bert}
Qian Chen, Zhu Zhuo, and Wen Wang. 2019. Bert for joint intent classification and slot filling. arXiv preprint:1902.10909.
\bibitem{deng2021htcinfomax}
Zhongfen Deng, Hao Peng, Dongxiao He, Jianxin Li, and Philip S. Yu. 2021. Htcinfomax: A global model for hierarchical text classification via information maximization. In Proceedings of the 2021 NAACL:HLT, pages 3259–3265.
\bibitem{deng-etal-2020-hierarchical}
Zhongfen Deng, Hao Peng, Congying Xia, Jianxin Li, Lifang He, and Philip Yu. 2020. Hierarchical bi-directional self-attention networks for paper review rating recommendation. In Proceedings of the 28th COLING, pages 6302–6314.
\bibitem{devlin2018bert}
Jacob Devlin, Ming-Wei Chang, Kenton Lee, and Kristina Toutanova. 2019. Bert: Pre-training of deep bidirectional transformers for language understanding. In Proceedings of the NAACL, pages 4171–4186.
\bibitem{Embar2021DiffXtractJD}
Varun R. Embar, Andrey Kan, Bunyamin Sisman, Christos Faloutsos, and Lise Getoor. 2021. DiffXtract: Joint Discriminative Product Attribute-Value Extraction. In Proceedings of ICBK, pages 271–280.
\bibitem{goo-etal-2018-slot}
Chih-Wen Goo, Guang Gao, Yun-Kai Hsu, Chih-Li Huo, Tsung-Chieh Chen, Keng-Wei Hsu, and Yun-Nung Chen. 2018. Slot-Gated Modeling for Joint Slot Filling and Intent Prediction. In Proceedings of the 2018 NAACL:HLT, pages 753–757. 

\bibitem{gopalakrishnan2012matching}
Vishrawas Gopalakrishnan, Suresh Parthasarathy Iyengar, Amit Madaan, Rajeev Rastogi, and Srinivasan Sengamedu. 2012. Matching product titles using web-based enrichment. In Proceedings of the 21st ACM CIKM, pages 605–614.
\bibitem{hakkanitur16_interspeech}
Dilek Hakkani-Tür, Gokhan Tur, Asli Celikyilmaz, Yun-Nung Chen, Jianfeng Gao, Li Deng, and Ye-Yi Wang. 2016. Multi-Domain Joint Semantic Frame Parsing Using Bi-Directional RNN-LSTM. In Proc. Interspeech 2016, pages 715–719.
\bibitem{han2019unsupervised}
Xiaochuang Han and Jacob Eisenstein. 2019. Unsupervised Domain Adaptation of Contextualized Embeddings for Sequence Labeling. In Proceedings of the 2019 EMNLP-IJCNLP, pages 4238–4248.


\bibitem{huang2019hierarchical}
Wei Huang, Enhong Chen, Qi Liu, Yuying Chen, Zai Huang, Yang Liu, Zhou Zhao, Dan Zhang, and Shijin Wang. 2019. Hierarchical multi-label text classification: An attention-based recurrent network approach. In Proceedings of the 28th ACM CIKM, pages 1051–1060.
\bibitem{johnson2015effective}
Rie Johnson and Tong Zhang. 2015. Effective use of word order for text categorization with convolutional neural networks. In Proceedings of the 2015 NAACL:HLT, pages 103–112.
\bibitem{karamanolakis2020txtract}
Giannis Karamanolakis, Jun Ma, and Xin Luna Dong. 2020. Txtract: Taxonomy-aware knowledge extraction for thousands of product categories. In Proceedings of the 58th ACL, pages 8489–8502.

\bibitem{liu16c_interspeech}
Bing Liu and Ian Lane. 2016. Attention-Based Recurrent Neural Network Models for Joint Intent Detection and Slot Filling. In Proc. Interspeech 2016, pages 685–689.
\bibitem{mao2019hierarchical}
Yuning Mao, Jingjing Tian, Jiawei Han, and Xiang Ren. 2019. Hierarchical text classification with reinforced label assignment. In Proceedings of the 2019 EMNLP-IJCNLP, pages 445–455.
\bibitem{mehta2021latex}
Kartik Mehta, Ioana Oprea, and Nikhil Rasiwasia. 2021. Latex-numeric: Language agnostic text attribute extraction for numeric attributes. In Proceedings of the 2021 NAACL:HLT: Industry Papers, pages 272–279.
\bibitem{more2016attribute}
Ajinkya More. 2016. Attribute extraction from product titles in ecommerce. arXiv preprint arXiv:1608.04670.
\bibitem{ni2019justifying}
Jianmo Ni, Jiacheng Li, and Julian McAuley. 2019. Justifying recommendations using distantly-labeled reviews and fine-grained aspects. In Proceedings of the 2019 EMNLP-IJCNLP, pages 188–197.
\bibitem{peng2018large}
Hao Peng, Jianxin Li, Yu He, Yaopeng Liu, Mengjiao Bao, Lihong Wang, Yangqiu Song, and Qiang Yang. 2018. Large-scale hierarchical text classification with recursively regularized deep graph-cnn. In Proceedings of the 2018 World Wide Web Conference, pages 1063–1072.
\bibitem{peng2019hierarchical}
Hao Peng, Jianxin Li, Senzhang Wang, Lihong Wang, Qiran Gong, Renyu Yang, Bo Li, Philip Yu, and Lifang He. 2019. Hierarchical taxonomy-aware and attentional graph capsule rcnns for large-scale multi-label text classification. IEEE TKDE, 33(6): 2505-2519.
\bibitem{radford2019language}
Alec Radford, Jeffrey Wu, Rewon Child, David Luan, Dario Amodei, Ilya Sutskever, et al. 2019. Language models are unsupervised multitask learners. OpenAI blog 1, 8 (2019), 9.
\bibitem{raffel2020exploring}
Colin Raffel, Noam Shazeer, Adam Roberts, Katherine Lee, Sharan Narang, Michael Matena, Yanqi Zhou, Wei Li, and Peter J Liu. 2020. Exploring the Limits of Transfer Learning with a Unified Text-to-Text Transformer. Journal of
Machine Learning Research 21 (2020), 1–67.
\bibitem{rajpurkar2016squad}
Pranav Rajpurkar, Jian Zhang, Konstantin Lopyrev, and Percy Liang. 2016. SQuAD: 100,000+ Questions for Machine Comprehension of Text. In Proceedings of the 2016 EMNLP, pages 2383–2392.
\bibitem{ramos2019attribute}
Kevin Almeida Ramos. 2019. Attribute-value inference using deep neural networks. Ph. D. Dissertation.
\bibitem{roy2021attribute}
Kalyani Roy, Pawan Goyal, and Manish Pandey. 2021. Attribute value generation from product title using language models. In Proceedings of The 4th Workshop on e-Commerce and NLP, pages 13–17.
\bibitem{see2017get}
Abigail See, Peter J Liu, and Christopher D Manning. 2017. Get to the point: Summarization with pointer-generator networks. arXiv preprint arXiv:1704.04368.
\bibitem{sharma2016automatic}
Vasu Sharma and Harish Karnick. 2016. Automatic tagging and retrieval of E-Commerce products based on visual features. In Proceedings of the NAACL Student Research Workshop, pages 22–28.

\bibitem{vandic2012faceted}
Damir Vandic, Jan-Willem Van Dam, and Flavius Frasincar. 2012. Faceted product search powered by the semantic web. Decision Support Systems, 53(3):425–437.
\bibitem{wang2020learning}
Qifan Wang, Li Yang, Bhargav Kanagal, Sumit Sanghai, D Sivakumar, Bin Shu, Zac Yu, and Jon Elsas. 2020. Learning to extract attribute value from product via question answering: A multi-task approach. In Proceedings of the 26th ACM SIGKDD, pages 47–55.
\bibitem{wang-etal-2022-smartave}
Qifan Wang, Li Yang, Jingang Wang, Jitin Krishnan, Bo Dai, Sinong Wang, Zenglin Xu, Madian Khabsa, and Hao Ma. 2022. SMARTAVE: Structured Multimodal Transformer for Product Attribute Value Extraction. In Findings of EMNLP 2022, pages 263–276.

\bibitem{wehrmann2018hierarchical}
Jonatas Wehrmann, Ricardo Cerri, and Rodrigo Barros. 2018. Hierarchical multi-label classification networks. In ICML, pages 5075–5084. 
\bibitem{wu-etal-2019-transferable}
Chien-Sheng Wu, Andrea Madotto, Ehsan Hosseini-Asl, Caiming Xiong, Richard Socher, and Pascale Fung. 2019. Transferable Multi-Domain State Generator for Task-Oriented Dialogue Systems. In Proceedings of the 57th ACL, pages 808–819.
\bibitem{wu2019learning}
Jiawei Wu, Wenhan Xiong, and William Yang Wang. 2019. Learning to learn and predict: A meta-learning approach for multi-label classification. In Proceedings of the 2019 EMNLP-IJCNLP, pages 4354–4364.
\bibitem{xu-etal-2019-scaling}
Huimin Xu, Wenting Wang, Xinnian Mao, Xinyu Jiang, and Man Lan. 2019. Scaling up open tagging from tens to thousands: Comprehension empowered attribute value extraction from product title. In Proceedings of the 57th ACL, pages 5214–5223.
\bibitem{yan2021adatag}
Jun Yan, Nasser Zalmout, Yan Liang, Christan Grant, Xiang Ren, and Xin Luna Dong. 2021. Adatag: Multi-attribute value extraction from product profiles with adaptive decoding. arXiv preprint arXiv:2106.02318.
\bibitem{yang2022mave}
Li Yang, Qifan Wang, Zac Yu, Anand Kulkarni, Sumit Sanghai, Bin Shu, Jon Elsas, and Bhargav Kanagal. 2022. Mave: A product dataset for multi-source attribute value extraction. In Proceedings of the Fifteenth ACM WSDM, pages 1256–1265.
\bibitem{you2019attentionxml}
Ronghui You, Zihan Zhang, Ziye Wang, Suyang Dai, Hiroshi Mamitsuka, and Shanfeng Zhu. 2019. Attentionxml: Label tree-based attention-aware deep model for high-performance extreme multi-label text classification. In NeurIPS, pages 5820–5830.
\bibitem{zhang-etal-2021-pdaln}
Tao Zhang, Congying Xia, Philip S. Yu, Zhiwei Liu, and Shu Zhao. 2021. PDALN: Progressive Domain Adaptation over a Pre-trained Model for Low-Resource Cross-Domain Named Entity Recognition. In Proceedings of the 2021 EMNLP, pages 5441–5451.
\bibitem{zheng2018opentag}
Guineng Zheng, Subhabrata Mukherjee, Xin Luna Dong, and Feifei Li. 2018. Opentag: Open attribute value extraction from product profiles. In Proceedings of the 24th ACM SIGKDD, pages 1049–1058.
\bibitem{zhou2020hierarchy}
Jie Zhou, Chunping Ma, Dingkun Long, Guangwei Xu, Ning Ding, Haoyu Zhang, Pengjun Xie, and Gong shen Liu. 2020. Hierarchy-aware global model for hierarchical text classification. In Proceedings of the 58th ACL, pages 1106–1117.
\bibitem{zhu-etal-2020-multimodal}
Tiangang Zhu, Yue Wang, Haoran Li, Youzheng Wu, Xiaodong He, and Bowen Zhou. 2020. Multimodal Joint Attribute Prediction and Value Extraction for E-commerce Product. In Proceedings of the 2020 EMNLP, pages 2129–2139.
\bibitem{deng2022aesmnsmlc}
Zhongfen Deng, Wei-Te Chen, Lei Chen, and Philip Yu. 2022. AE-smnsMLC: Multi-Label Classification with Semantic Matching and Negative Label Sampling for Product Attribute Value Extraction. In Proceedings of 2022 Big Data, pages 1816-1821.

\end{thebibliography}
\end{document}